\pdfoutput=1

\documentclass[11pt]{article}

\usepackage[final]{acl}
\usepackage{booktabs}
\usepackage{times}
\usepackage{latexsym}

\usepackage[main=english, russian]{babel}
\addto\captionsenglish{}
\usepackage{CJKutf8}
\usepackage{graphicx}
\usepackage{subcaption}
\usepackage[T1]{fontenc}

\usepackage[utf8]{inputenc}

\usepackage{microtype}

\usepackage{inconsolata}

\title{AILS-NTUA at SemEval-2024 Task 6: Efficient model tuning for hallucination detection and analysis}

\author{
    Natalia Grigoriadou,  Maria Lymperaiou, Giorgos Filandrianos, Giorgos Stamou\\ 
    School of Electrical and Computer Engineering,  AILS Laboratory\\
    National Technical University of Athens \\
      \texttt{\href{mailto:natalygrigoriadi@gmail.com}{natalygrigoriadi@gmail.com}, \{\href{mailto:marialymp@islab.ntua.gr}{marialymp}, \href{mailto:geofila@islab.ntua.gr}{geofila}}\}@islab.ntua.gr \\
  \texttt{\href{mailto:gstam@cs.ntua.gr}{gstam@cs.ntua.gr}}
}

\begin{document}
\maketitle
\begin{abstract}
In this paper, we present our team's submissions for SemEval-2024 Task-6 - SHROOM, a Shared-task on Hallucinations and Related Observable Overgeneration Mistakes. The participants were asked to perform binary classification to identify cases of fluent overgeneration hallucinations. Our experimentation included fine-tuning a pre-trained model on hallucination detection and a Natural Language Inference (NLI) model. The most successful strategy involved creating an ensemble of these models, resulting in accuracy rates of 77.8\% and 79.9\% on model-agnostic and model-aware datasets respectively, outperforming the organizers' baseline and achieving notable results when contrasted with the top-performing results in the competition, which reported accuracies of 84.7\% and 81.3\% correspondingly.
\end{abstract}

\section{Introduction}
In the era that Large Language Models (LLMs) dominate and shape the trends in the Natural Language Processing (NLP) community, ensuring reliance and accurate functionality of related systems becomes a major concern. Hallucinations of language models have recently received lots of attention \cite{rawte2023survey, survey-hal, huang2023survey, ye2023cognitive, Zhang2023SirensSI}, questioning the trust that humans can pose in highly intelligent yet probabilistic models. At the same time, recent endeavors formally prove that hallucinations are inherent to LLMs and thus inevitable in practice \cite{Xu2024HallucinationII}. 

Encompassing the need for detecting and analyzing hallucinations in Natural Language Generation (NLG) tasks, and given the scarcity of related datasets and benchmarks \cite{li-etal-2023-halueval, Cao2023AutoHallAH, Chen2023FactCHDBF, muhlgay2024generating}, the SemEval-2024 Task 6 (SHROOM: a Shared-task on Hallucinations and Related Observable Overgeneration Mistakes) \cite{mickus-etal-2024-semeval} addresses the presence of semantically unrelated generations with respect to a given input, covering challenging NLP tasks, such as Machine Translation, Definition Modelling and Paraphrase Generation, which are tested both when the underlying model is known or not.

To this end, we explore efficient and widely adaptable hallucination detection strategies, tailored to the black-box demands of the problem\footnote{Even in the model-aware setting of SHROOM, we do not re-generate the outputs using the given models, therefore we continue operating in a completely black-box setup.}. Based on pre-trained models which contain knowledge regarding semantic relationships related to hallucinations, we achieve $\sim$80\% accuracy in hallucination detection by fine-tuning on labeled SHROOM instances, notably higher than the 74.5\% baseline accuracy provided, using an open-source Mistral instruction-tuned model\footnote{\url{https://huggingface.co/TheBloke/Mistral-7B-Instruct-v0.2-GGUF}}. Specifically, we contribute to the following: 
\begin{enumerate}
    \item We fine-tune models pre-trained on hallucination detection and Natural Language Inference (NLI) datasets, which are semantically related to SHROOM challenges.
    \item Tuned models constitute a Voting Classifier, achieving competitive detection accuracy.
    \item All our experimentation is time and computationally efficient, while entirely black-box.
    \item Decomposition of results per task and analysis of failed and accurately detected instances provide valuable insights into the nature of the involved hallucinations.
\end{enumerate}

Our code is available on GitHub \footnote{\url{https://github.com/ngregoriade/Semeval2024-Shroom.git}}.

\section{Related Work}
\paragraph{NLP hallucinations} is a rapidly evolving field, examining invalid generations from varying perspectives. Categorizations of hallucinations may view hallucinatory outputs as unfaithful to the input, inconsistent with the generated output itself, or conflicting with real-world knowledge \cite{Zhang2023SirensSI}. Factual hallucinations have gathered the majority of recent breakthroughs, since comparison with existing factual sources \cite{lin2022truthfulqa, lee2023factuality, Chen2023FactCHDBF, min2023factscore, Cao2023AutoHallAH, muhlgay2024generating} renders them accurately detectable and correctable \cite{chern2023factool, dhuliawala2023chainofverification, li2024dawn}. The more subtle characteristics of other hallucination types constitute the creation of related benchmarks harder, not to mention techniques for automatic evaluation \cite{azaria2023internal, kadavath2022language, manakul2023selfcheckgpt, Duan2024DoLK}. A limitation tied with such techniques is that in most cases at least model probing is needed, rendering them unusable in cases where the model that produced the reported hallucinations is completely unknown or inaccessible.
SHROOM comes to fill this gap, focusing on semantic faithfulness rather than factuality, while requesting a diverging suite of proposed detection techniques that should even cover cases that the model is not given at all. As a trade-off, implementations on the SHROOM dataset require the ground-truth output, since the given input does not contain the necessary semantic information to drive decisions on whether a sample is a hallucination or not.
Our proposed approach only considers given \textit{inputs} and \textit{outputs} and does \textit{not} probe any model, contrary to other black-box techniques \cite{manakul2023selfcheckgpt}.

\section{Task and Dataset description}
Driven by upcoming challenges in the NLG landscape, SHROOM dataset focuses on the prevalent issues of models generating linguistically fluent but inaccurate (incorrect or unsupported) outputs.  Participants are tasked with binary classification to identify instances of fluent overgeneration hallucinations in \textit{model-aware} and \textit{model-agnostic} tracks. The task encompasses three NLG domains—definition modeling (DM), machine translation (MT), and paraphrase generation (PG)—with provided checkpoints, inputs, references, and outputs for binary classification. The development set includes annotations from multiple annotators, establishing a majority vote gold label.

\paragraph{Data details} In all cases, data follow a specific format: \textit{src} is the input given to a model, \textit{hyp} is the output generated by the model, \textit{tgt} comprises the ground truth output for this specific model, \textit{ref} indicates whether target, source or both of these fields contain the semantic information necessary to establish whether a datapoint is a hallucination,
\textit{task} refers to the task being solved and \textit{model} to the model being used (in the model-agnostic case the \textit{model} entry remains empty). An example of the data format is given in Table \ref{tab:examples-unlabelled}. Initially, 80 labeled trial samples were released, followed by unlabelled training data which contain 30k model-agnostic and 30k model-aware instances. Finally, the labeled validation set contains 499 and 501 samples for model-agnostic and model-aware settings respectively, while
the test set comprises 1500 model-agnostic and 1500 model-aware labeled samples. Additional information provided in the labeled splits are \textit{labels}, which contains a list of `Hallucination' and `Not Hallucination' labels as provided by 5 annotators per sample, the final \textit{label} occurring via majority voting over the aforementioned list and \textit{p(Hallucination)}, denoting the probability of hallucination as the percentage of agreeing annotators on the `Hallucination' label.
A thorough data analysis is provided in the App. \ref{sec:analysis}.

\paragraph{Evaluation metrics} proposed from the task organizers for SHROOM are accuracy, regarding the classification success in `Hallucination'/`Not Hallucination' classes and Spearman correlation (RHO), measuring the -positive- correlation between validation and test p(`Hallucination') values.

\section{Methods}
As the core of our system, we propose a universal and lightweight methodology that leverages well-established pre-trained classifiers for hallucination detection.  We propose 3 techniques to approach it.
\subsection{Fine-tune hallucination detection model}
\label{sec:hal-model}
Our first technique employs fine-tuning a pre-trained classifier dedicated to hallucination detection to learn distinguishing patterns between hallucinated/non-hallucinated SHROOM instances. 
More specifically, we employed a pre-trained model based on microsoft/deberta-v3-base provided by Hugging Face\footnote{\url{https://huggingface.co/vectara/hallucination_evaluation_model}}, especially designed for hallucination detection. This model was initially trained on NLI data to ascertain textual entailment. Subsequently, it underwent further fine-tuning using summarization datasets enriched with factual consistency annotations. The output of our employed model is a probability score in the [0, 1] range; a score of 0 indicates the presence of hallucination in the generated content, while a score of 1 signifies factual consistency. This probabilistic nature enables the evaluation of the model's confidence in the veracity of the generated hypotheses.

To tailor the model to the specific demands of our task, we used the provided annotated validation set of 1000 samples for training purposes. This adaptation process aimed to enhance the model's performance by aligning it with the variation and complexity present in SHROOM.
Moreover, we applied a thresholding approach to make practical decisions based on the probabilistic outputs of the model.
By setting a threshold at 0.5, we categorize predictions with scores above this threshold as indicative of input-output consistency, while the rest are considered as potential hallucinatory instances. 

\subsection{Fine-tune NLI models}
\label{sec:nli}
In the context of detecting hallucinated answers, we also employed NLI models, an approach that has witnessed significant advancements, while investigating semantic intricacies close to hallucinations. NLI models play a crucial role in enabling comprehension of the sophisticated connections between sentences, categorizing the relationship between a \textit{hypothesis} and a \textit{premise} into entailment, neutral, or contradiction.
In terms of our task, we convert hallucination detection to an NLI problem: given the input (termed as \textit{hypothesis-hyp}) to a model and the premise (named \textit{target-tgt}) we evaluate whether \textit{tgt} entails, contradicts or remains neutral to \textit{hyp}.

To execute this approach in technical terms, we select a pre-trained NLI model available through Hugging Face\footnote{\url{https://huggingface.co/MoritzLaurer/mDeBERTa-v3-base-xnli-multilingual-nli-2mil7}}. This model, based on mDeBERTa-v3-base architecture, was originally trained on a large-scale multilingual dataset, making it well-suited for handling diverse linguistic details.
To fine-tune the NLI model and tailor it to the specific intricacies of our task, we employed the annotated validation set, as in the previous case. 

\subsection{Voting Classifier}
In our final approach, we employed an ensemble technique known as a Voting Classifier.  The underlying principle is to aggregate the collective insights derived from each constituent classifier (in our case the previously mentioned models), ultimately predicting the output class based on the highest majority of votes.
By doing so, the ensemble not only leverages the individual strengths of each method but also mitigates potential weaknesses, thereby enhancing the overall predictive performance in a deliberate effort to address the inherent complexity and variability within the dataset, contributing to a more nuanced and accurate understanding of the phenomena under investigation.

\section{Experiments}
\subsection{Experimental setup}
All our experiments were executed using Google Colab platform with a single Tesla T4 GPU.
\paragraph{Fine-tune hallucination model}
Our fine-tuned model underwent a rigorous training and evaluation process, utilizing SHROOM data provided by the task organizers. Specifically, the model was trained with the annotated validation set and evaluated against the trial set.
In the pre-processing phase, from each data point, we extracted the \textit{hyp} and \textit{tgt} components to serve as inputs to the model. 

To optimize the model's performance in terms of both accuracy and p(`Hallucination'), we implemented a dual-training strategy. The model was trained twice, employing binary labels (0 for Hallucination and 1 for Not Hallucination) in one iteration and float labels (representing 1-p(`Hallucination')) in the other. This dual-training approach allowed us to derive two crucial aspects from the model: the binary label indicating the presence or absence of hallucination, and the corresponding probability score indicating the likelihood of hallucination.
The hyperparameters for fine-tuning are comprehensively detailed in Table~\ref{Parameters used for fine-tuning the pre-trained model}.

\begin{table}[h!]
\centering
\small
\begin{tabular}{ll}
\hline
\textbf{Hyperparameter} & \textbf{Value} \\
\hline
train dataloader & validation set (1,000 samples)  \\
evaluator & trial set (80 samples) \\
epochs & 5 \\
evaluation steps & 10,000 \\
warm-up steps & 10\% of train data for warm-up \\
\hline
\end{tabular}
\caption{\label{Parameters used for fine-tuning the pre-trained model}
Hyperparameters used for the hallucination detection model fine-tuning}
\end{table}

\paragraph{Natural Language Inference (NLI) models}

This NLI model was already trained with the multilingual-nli-26lang-2mil7 \cite{laurer_less_2022} dataset and the XNLI  validation dataset \cite{conneau2018xnli}, both containing three different labels: `entailment', `neutral' and `contradiction'. During the training phase, we systematically mapped the `Hallucination' label to `contradiction' and the `Not Hallucination' label to `entailment', ensuring a binary representation of the hallucinatory nature of the content. This transformation facilitated the training process by providing clear labels for the model to learn the distinctions between hallucinatory and non-hallucinatory instances.
  
Post-training, the model's predictions were assessed using the entailment score, and a strategically chosen threshold was employed to distinguish between hallucinations and non-hallucinations. Prior to training, we experimented with a wide range of threshold values, concluding that a threshold of 0.8 optimized the accuracy of the trial set. Simultaneously, for the determination of the percentage of Hallucination for each data point, we used the entailment percentage subtracted from 1.

A detailed account of the parameters employed for training this NLI model is outlined in Table~\ref{Nli parameters}. 
\begin{table}[h!]
\small
\centering
\begin{tabular}{ll}
\hline
\textbf{Hyperparameter} & \textbf{Value} \\
\hline
train dataset & validation set (1,000 samples)  \\
learning rate & 2e-05\\

epochs & 5 \\
warm-up ratio & 0.06 \\
weight decay & 0.01\\

\hline
\end{tabular}
\caption{\label{Nli parameters}
Hyperparameters used for NLI fine-tuning
}
\end{table}

\paragraph{Voting Classifier} \label{Voting Classifier}
In the final leg of our methodological exploration, the Voting Classifier integrates the pre-trained hallucination detection model, its fine-tuned counterpart from \S\ref{sec:hal-model}, and the fine-tuned NLI model described in \S\ref{sec:nli}. 

The Voting Classifier operates on a dual strategy for hallucination categorization. First, for the binary labels, we assigned the majority label (`Hallucination' or 'Not Hallucination') among the three models to each data point.
Second, to determine the percentage of hallucination for each data point, we provided two methodologies. For the first one, we implemented a similar methodology to the one used in the validation and trial sets, i.e. the percentage of hallucination derived from the majority vote of the annotators. By emulating the same process, we calculate the percentage of models that labeled a given data point as `Hallucination'.  For the second one, we use the float p(`Hallucination') scores of each of the three models constituting the ensemble and extract the average value.

\subsection{Results}
\paragraph{Baseline System} During the evaluation phase, we were provided with a baseline system, which was based on a simple prompt retrieval approach, derived from SelfCheck-GPT\cite{manakul2023selfcheckgpt}, using an open-source Mistral instruction-tuned model as its core component (the prompt is shown in Table \ref{Prompt Baseline}). If the answer starts with`Yes' the sample is classified as `Not Hallucination' with p('Hallucination') equal to the probability that the token was chosen subtracted from 1, else if the answer starts with `No' the sample is classified as `Hallucination' with p(`Hallucination') equal to the probability that the token was chosen. If the answer starts with neither, the label is assigned randomly and p(`Hallucination') equals to 0.5.

\paragraph{Averaged results} for all our experiments are presented in Table~\ref{Final results}. The Voting Classifier achieves top results, with a more notable difference in the model-agnostic setting. This is an expected behavior since the ensembling of models is designed to boost the performance of its standalone constituents.
\begin{table}[h!]
\small
\centering
\begin{tabular}{p{4.8cm}cc}
\hline
\textbf{Method} & \textbf{acc.}$\uparrow$ & \textbf{rho}$\uparrow$ \\
\hline
\multicolumn{3}{c}{\textbf{Model-aware}} \\
\hline
Baseline Model & 0.745 & 0.488\\
Fine-tune hal-detect model & 0.795 & 0.685  \\
NLI model & 0.77 & 0.591 \\
Voting Classifier-majority vote & \textbf{0.799}& 0.691  \\
Voting Classifier-averaged percentage&\textbf{0.799}&\textbf{0.693} \\
\hline
\multicolumn{3}{c}{\textbf{Model-agnostic}} \\
\hline
Baseline Model & 0.697 &  0.402\\
Fine-tune hal-detect model & 0.778 & 0.668 \\
NLI model & 0.751 & 0.548\\
Voting Classifier-majority vote & \textbf{0.78} & 0.632\\
Voting Classifier-averaged percentage & \textbf{0.78}&\textbf{0.643}\\
\hline
\end{tabular}
\caption{\label{Final results}
Final results for model-aware and model-agnostic variants. \textbf{Bold} denotes best results. The two Voting Classifiers differentiate from the method applied to calculate the p(`Hallucination) as explained in \ref{Voting Classifier}}
\label{tab:accents}
\end{table}
\begin{figure*}[h!]
    \centering
    \includegraphics[width=0.8\textwidth]{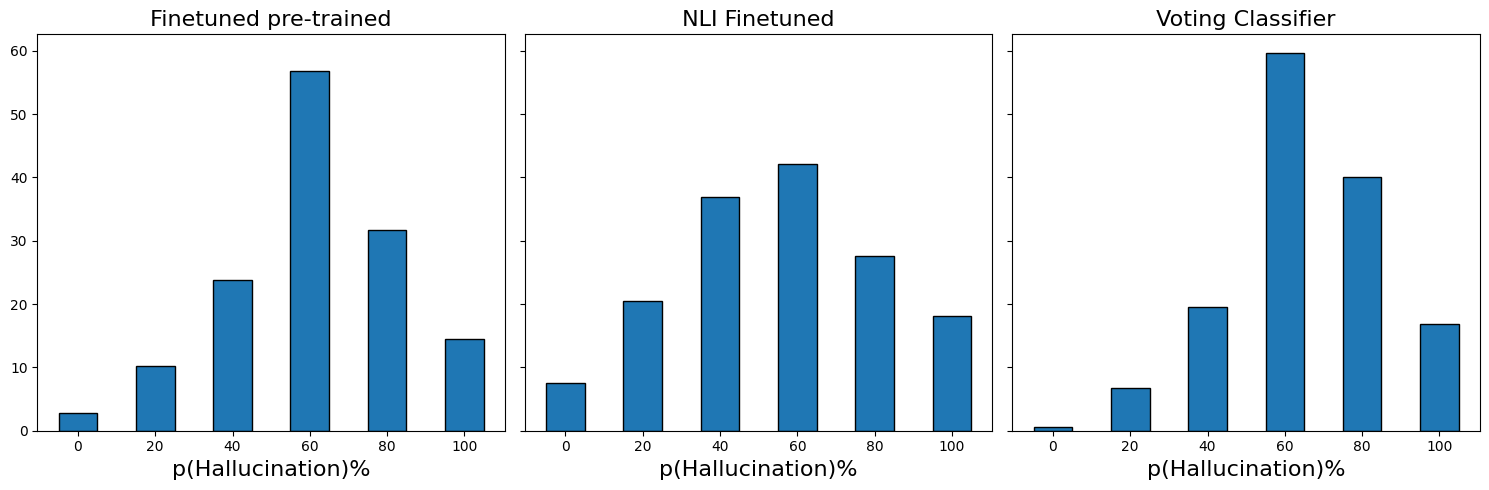}
    \caption{p(`Hallucination') for all misclassified samples of model aware dataset.}
    \label{fig:error-phal-aware}
\end{figure*}
\begin{figure*}[h!]
\vskip -5px
    \centering
    \includegraphics[width=0.8\textwidth]{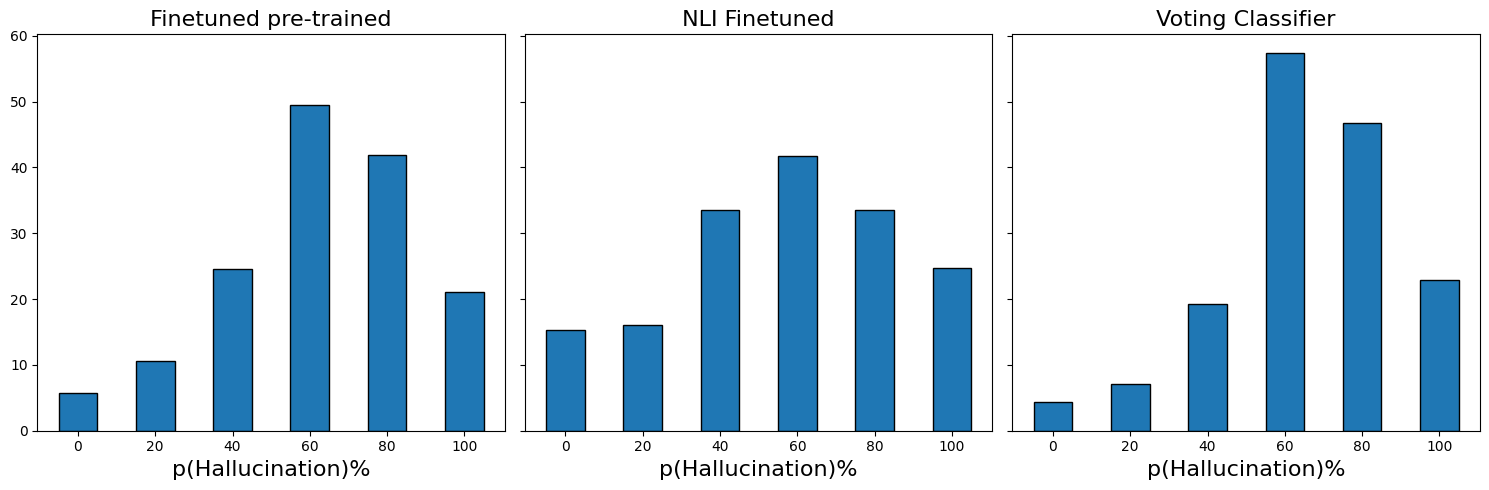}
    \caption{p(`Hallucination') for all misclassified samples of model agnostic dataset.}
    \label{fig:error-phal-agnostic}
\end{figure*}

We demonstrate the computational efficiency of our proposed methods regarding the training and inference time needed in Table \ref{tab:time}. The Voting Classifier sums the times of all three of its model-voters. Since reported runtimes were achieved using the T4 GPU of the free Google Colab version, our proposed methods can be replicated and utilized by any user, without any budget or time limitations, nor the need to access sophisticated hardware.
\begin{table}[h!]
\small
\centering
\begin{tabular}{lcccc}
\hline
\textbf{Method} & \textbf{Training}$\downarrow$ & \textbf{Inference}$\downarrow$\\
\hline
pre-trained hal-detect model & - & 39.00\\
Fine-tune hal-detect model & 91.59 & 45.66\\
NLI model & 927.14  & 58.96\\
Voting Classifier & 1,018.73 & 143.62\\
\hline
\end{tabular}
\caption{\label{Times}
Training and inference time in seconds.}
\label{tab:time}
\end{table}

\begin{table}[h!]
\centering
\small
\begin{tabular}{lp{0.3cm}c}
\hline
\textbf{NLG Model} & \textbf{Task} & \textbf{aware-acc}$\uparrow$\\
\hline
\multicolumn{3}{c}{\textbf{Hal-detect model fine-tuning}} \\
\hline
tuner007/pegasus\_paraphrase & PG & 0.856 \\
facebook/nllb-200-distilled-600M& MT & 0.824\\
ltg/flan-t5-definition-en-base & DM & 0.724\\
\hline
\multicolumn{3}{c}{\textbf{NLI model fine-tuning}} \\
\hline
tuner007/pegasus\_paraphrase & PG & 0.803 \\
facebook/nllb-200-distilled-600M& MT & 0.789\\
ltg/flan-t5-definition-en-base & DM & 0.703\\
\hline
\multicolumn{3}{c}{\textbf{Voting Classifier}} \\
\hline
tuner007/pegasus\_paraphrase & PG & 0.861 \\
facebook/nllb-200-distilled-600M& MT & 0.828\\
ltg/flan-t5-definition-en-base & DM & 0.73\\
\hline
\end{tabular}
\caption{\label{tab:aware}
Model-aware accuracy per model and task.}
\end{table}
 
Moreover, per-task and model hallucination detection for the model-aware dataset is presented in Table \ref{tab:aware}. The PG task demonstrates superior performance compared to the other two tasks, while the DM task reports significantly lower accuracy. This disparity in outcomes can be explained by the inherent characteristics of each task when formulated as a paraphrase problem. The PG task exhibits notably higher results owing to its direct alignment with the paraphrase objective. Similarly, the MT task, which evaluates translations from the LLM against ground truth translation, achieves relatively comparable results. Conversely, the DM task faces the complexities of articulating precise and contextually relevant definitions. Consequently, the DM task exhibits notably lower accuracy due to the intricacies of handling more complex sentence structures. The Voting Classifier remains the top scorer in each of the tasks, highlighting the power of ensembling individual predictors.  

Finally, we perform \textit{some error} analysis on the misclassified samples (Figures \ref{fig:error-phal-aware}, \ref{fig:error-phal-agnostic}): we measure the p(`Hallucination') for misclassifications for all our 3 methods. Ideally, p(`Hallucination') values for misclassifications should lie close to the discrimination threshold of 0.5, indicating that their separability is highly uncertain. Indeed, our best performing Voting Classifier presents a peak for p(`Hallucination')=0.6 for both model-aware and model-agnostic settings, highlighting that misclassified samples are in any case hard to classify in their correct class. Moreover, the p(`Hallucination') values in the range [0.0-0.4] -corresponding to the `Not Hallucination' label-  are lower for the Voting Classifier in comparison to the other two models, denoting that ensembling reduces misclassifications for non-hallucinatory instances.

\section{Conclusion}
In this work, we detect and analyze hallucinations from the SHROOM dataset introduced in SemEval 2024 Task 6. We propose a computationally efficient methodology based on fine-tuning models that present semantic cues close to SHROOM's hallucinations, while model ensembling further boosts results in 3 NLG tasks. Our techniques operate in a fully black-box setting, solely requiring inputs and outputs obtained from NLG models. Our error analysis demonstrates that our misclassifications are samples of high uncertainty in terms of hallucination probability and, therefore hard to be discerned overall. In total, we aspire that our simple though efficient technique will assist future research in the crucial hallucination detection field.
    


\bibliography{acl_latex}

\begin{thebibliography}{23}
\expandafter\ifx\csname natexlab\endcsname\relax\def\natexlab#1{#1}\fi

\bibitem[{Azaria and Mitchell(2023)}]{azaria2023internal}
Amos Azaria and Tom Mitchell. 2023.
\newblock \href {http://arxiv.org/abs/2304.13734} {The internal state of an llm knows when it's lying}.

\bibitem[{Cao et~al.(2023)Cao, Yang, and Zhao}]{Cao2023AutoHallAH}
Zouying Cao, Yifei Yang, and Hai Zhao. 2023.
\newblock \href {https://api.semanticscholar.org/CorpusID:263334406} {Autohall: Automated hallucination dataset generation for large language models}.
\newblock \emph{ArXiv}, abs/2310.00259.

\bibitem[{Chen et~al.(2023)Chen, Song, Gui, Wang, Zhang, Yong, Huang, Lv, Zhang, and Chen}]{Chen2023FactCHDBF}
Xiang Chen, Duanzheng Song, Honghao Gui, Chenxi Wang, Ningyu Zhang, Jiang Yong, Fei Huang, Chengfei Lv, Dan Zhang, and Huajun Chen. 2023.
\newblock \href {https://api.semanticscholar.org/CorpusID:264289140} {Factchd: Benchmarking fact-conflicting hallucination detection}.
\newblock \emph{ArXiv}, abs/2310.12086.

\bibitem[{Chern et~al.(2023)Chern, Chern, Chen, Yuan, Feng, Zhou, He, Neubig, and Liu}]{chern2023factool}
I-Chun Chern, Steffi Chern, Shiqi Chen, Weizhe Yuan, Kehua Feng, Chunting Zhou, Junxian He, Graham Neubig, and Pengfei Liu. 2023.
\newblock \href {http://arxiv.org/abs/2307.13528} {Factool: Factuality detection in generative ai -- a tool augmented framework for multi-task and multi-domain scenarios}.

\bibitem[{Conneau et~al.(2018)Conneau, Rinott, Lample, Williams, Bowman, Schwenk, and Stoyanov}]{conneau2018xnli}
Alexis Conneau, Ruty Rinott, Guillaume Lample, Adina Williams, Samuel~R. Bowman, Holger Schwenk, and Veselin Stoyanov. 2018.
\newblock Xnli: Evaluating cross-lingual sentence representations.
\newblock In \emph{Proceedings of the 2018 Conference on Empirical Methods in Natural Language Processing}. Association for Computational Linguistics.

\bibitem[{Dhuliawala et~al.(2023)Dhuliawala, Komeili, Xu, Raileanu, Li, Celikyilmaz, and Weston}]{dhuliawala2023chainofverification}
Shehzaad Dhuliawala, Mojtaba Komeili, Jing Xu, Roberta Raileanu, Xian Li, Asli Celikyilmaz, and Jason Weston. 2023.
\newblock \href {http://arxiv.org/abs/2309.11495} {Chain-of-verification reduces hallucination in large language models}.

\bibitem[{Duan et~al.(2024)Duan, Yang, and Tam}]{Duan2024DoLK}
Hanyu Duan, Yi~Yang, and Kar~Yan Tam. 2024.
\newblock \href {https://api.semanticscholar.org/CorpusID:267682191} {Do llms know about hallucination? an empirical investigation of llm's hidden states}.

\bibitem[{Huang et~al.(2023)Huang, Yu, Ma, Zhong, Feng, Wang, Chen, Peng, Feng, Qin, and Liu}]{huang2023survey}
Lei Huang, Weijiang Yu, Weitao Ma, Weihong Zhong, Zhangyin Feng, Haotian Wang, Qianglong Chen, Weihua Peng, Xiaocheng Feng, Bing Qin, and Ting Liu. 2023.
\newblock \href {http://arxiv.org/abs/2311.05232} {A survey on hallucination in large language models: Principles, taxonomy, challenges, and open questions}.

\bibitem[{Ji et~al.(2023)Ji, Lee, Frieske, Yu, Su, Xu, Ishii, Bang, Madotto, and Fung}]{survey-hal}
Ziwei Ji, Nayeon Lee, Rita Frieske, Tiezheng Yu, Dan Su, Yan Xu, Etsuko Ishii, Ye~Jin Bang, Andrea Madotto, and Pascale Fung. 2023.
\newblock \href {https://doi.org/10.1145/3571730} {Survey of hallucination in natural language generation}.
\newblock \emph{ACM Comput. Surv.}, 55(12).

\bibitem[{Kadavath et~al.(2022)Kadavath, Conerly, Askell, Henighan, Drain, Perez, Schiefer, Hatfield-Dodds, DasSarma, Tran-Johnson, Johnston, El-Showk, Jones, Elhage, Hume, Chen, Bai, Bowman, Fort, Ganguli, Hernandez, Jacobson, Kernion, Kravec, Lovitt, Ndousse, Olsson, Ringer, Amodei, Brown, Clark, Joseph, Mann, McCandlish, Olah, and Kaplan}]{kadavath2022language}
Saurav Kadavath, Tom Conerly, Amanda Askell, Tom Henighan, Dawn Drain, Ethan Perez, Nicholas Schiefer, Zac Hatfield-Dodds, Nova DasSarma, Eli Tran-Johnson, Scott Johnston, Sheer El-Showk, Andy Jones, Nelson Elhage, Tristan Hume, Anna Chen, Yuntao Bai, Sam Bowman, Stanislav Fort, Deep Ganguli, Danny Hernandez, Josh Jacobson, Jackson Kernion, Shauna Kravec, Liane Lovitt, Kamal Ndousse, Catherine Olsson, Sam Ringer, Dario Amodei, Tom Brown, Jack Clark, Nicholas Joseph, Ben Mann, Sam McCandlish, Chris Olah, and Jared Kaplan. 2022.
\newblock \href {http://arxiv.org/abs/2207.05221} {Language models (mostly) know what they know}.

\bibitem[{Laurer et~al.(2022)Laurer, Atteveldt, Casas, and Welbers}]{laurer_less_2022}
Moritz Laurer, Wouter~van Atteveldt, Andreu~Salleras Casas, and Kasper Welbers. 2022.
\newblock \href {https://osf.io/74b8k} {Less {Annotating}, {More} {Classifying} – {Addressing} the {Data} {Scarcity} {Issue} of {Supervised} {Machine} {Learning} with {Deep} {Transfer} {Learning} and {BERT} - {NLI}}.
\newblock \emph{Preprint}.
\newblock Publisher: Open Science Framework.

\bibitem[{Lee et~al.(2023)Lee, Ping, Xu, Patwary, Fung, Shoeybi, and Catanzaro}]{lee2023factuality}
Nayeon Lee, Wei Ping, Peng Xu, Mostofa Patwary, Pascale Fung, Mohammad Shoeybi, and Bryan Catanzaro. 2023.
\newblock \href {http://arxiv.org/abs/2206.04624} {Factuality enhanced language models for open-ended text generation}.

\bibitem[{Li et~al.(2024)Li, Chen, Ren, Cheng, Zhao, Nie, and Wen}]{li2024dawn}
Junyi Li, Jie Chen, Ruiyang Ren, Xiaoxue Cheng, Wayne~Xin Zhao, Jian-Yun Nie, and Ji-Rong Wen. 2024.
\newblock \href {http://arxiv.org/abs/2401.03205} {The dawn after the dark: An empirical study on factuality hallucination in large language models}.

\bibitem[{Li et~al.(2023)Li, Cheng, Zhao, Nie, and Wen}]{li-etal-2023-halueval}
Junyi Li, Xiaoxue Cheng, Xin Zhao, Jian-Yun Nie, and Ji-Rong Wen. 2023.
\newblock \href {https://doi.org/10.18653/v1/2023.emnlp-main.397} {{H}alu{E}val: A large-scale hallucination evaluation benchmark for large language models}.
\newblock In \emph{Proceedings of the 2023 Conference on Empirical Methods in Natural Language Processing}, pages 6449--6464, Singapore. Association for Computational Linguistics.

\bibitem[{Lin et~al.(2022)Lin, Hilton, and Evans}]{lin2022truthfulqa}
Stephanie Lin, Jacob Hilton, and Owain Evans. 2022.
\newblock \href {http://arxiv.org/abs/2109.07958} {Truthfulqa: Measuring how models mimic human falsehoods}.

\bibitem[{Manakul et~al.(2023)Manakul, Liusie, and Gales}]{manakul2023selfcheckgpt}
Potsawee Manakul, Adian Liusie, and Mark J.~F. Gales. 2023.
\newblock \href {http://arxiv.org/abs/2303.08896} {Selfcheckgpt: Zero-resource black-box hallucination detection for generative large language models}.

\bibitem[{Mickus et~al.(2024)Mickus, Zosa, V{\'a}zquez, Vahtola, Tiedemann, Segonne, Raganato, and Apidianaki}]{mickus-etal-2024-semeval}
Timothee Mickus, Elaine Zosa, Ra{\'u}l V{\'a}zquez, Teemu Vahtola, J{\"o}rg Tiedemann, Vincent Segonne, Alessandro Raganato, and Marianna Apidianaki. 2024.
\newblock {S}em{E}val-2024 {T}ask 6: {SHROOM}, a shared-task on hallucinations and related observable overgeneration mistakes.
\newblock In \emph{Proceedings of the 18th International Workshop on Semantic Evaluation (SemEval-2024)}, Mexico City, Mexico. Association for Computational Linguistics.

\bibitem[{Min et~al.(2023)Min, Krishna, Lyu, Lewis, tau Yih, Koh, Iyyer, Zettlemoyer, and Hajishirzi}]{min2023factscore}
Sewon Min, Kalpesh Krishna, Xinxi Lyu, Mike Lewis, Wen tau Yih, Pang~Wei Koh, Mohit Iyyer, Luke Zettlemoyer, and Hannaneh Hajishirzi. 2023.
\newblock \href {http://arxiv.org/abs/2305.14251} {Factscore: Fine-grained atomic evaluation of factual precision in long form text generation}.

\bibitem[{Muhlgay et~al.(2024)Muhlgay, Ram, Magar, Levine, Ratner, Belinkov, Abend, Leyton-Brown, Shashua, and Shoham}]{muhlgay2024generating}
Dor Muhlgay, Ori Ram, Inbal Magar, Yoav Levine, Nir Ratner, Yonatan Belinkov, Omri Abend, Kevin Leyton-Brown, Amnon Shashua, and Yoav Shoham. 2024.
\newblock \href {http://arxiv.org/abs/2307.06908} {Generating benchmarks for factuality evaluation of language models}.

\bibitem[{Rawte et~al.(2023)Rawte, Sheth, and Das}]{rawte2023survey}
Vipula Rawte, Amit Sheth, and Amitava Das. 2023.
\newblock \href {http://arxiv.org/abs/2309.05922} {A survey of hallucination in large foundation models}.

\bibitem[{Xu et~al.(2024)Xu, Jain, and Kankanhalli}]{Xu2024HallucinationII}
Ziwei Xu, Sanjay Jain, and Mohan~S. Kankanhalli. 2024.
\newblock \href {https://api.semanticscholar.org/CorpusID:267069207} {Hallucination is inevitable: An innate limitation of large language models}.
\newblock \emph{ArXiv}, abs/2401.11817.

\bibitem[{Ye et~al.(2023)Ye, Liu, Zhang, Hua, and Jia}]{ye2023cognitive}
Hongbin Ye, Tong Liu, Aijia Zhang, Wei Hua, and Weiqiang Jia. 2023.
\newblock \href {http://arxiv.org/abs/2309.06794} {Cognitive mirage: A review of hallucinations in large language models}.

\bibitem[{Zhang et~al.(2023)Zhang, Li, Cui, Cai, Liu, Fu, Huang, Zhao, Zhang, Chen, Wang, Luu, Bi, Shi, and Shi}]{Zhang2023SirensSI}
Yue Zhang, Yafu Li, Leyang Cui, Deng Cai, Lemao Liu, Tingchen Fu, Xinting Huang, Enbo Zhao, Yu~Zhang, Yulong Chen, Longyue Wang, Anh~Tuan Luu, Wei Bi, Freda Shi, and Shuming Shi. 2023.
\newblock \href {https://api.semanticscholar.org/CorpusID:261530162} {Siren's song in the ai ocean: A survey on hallucination in large language models}.
\newblock \emph{ArXiv}, abs/2309.01219.

\end{thebibliography}

\appendix

\section{Organizers' baseline}
The prompt used by the organizers to construct the baseline   Mistral instruction-tuned model is demonstrated in Table \ref{Prompt Baseline}.

\begin{table}[htb]
\centering
\small
\begin{tabular}{l}
\hline
\textbf{Prompt} \\
\hline
Context \{tgt\}\\
Sentence: \{hyp\}\\
Is the sentence supported by the context above?\\
Answer Yes or No:\\
\hline
\end{tabular}
\caption{\label{Prompt Baseline}
Prompt used in the Baselined System}
\end{table}

\section{Data format}
In Table \ref{tab:examples-unlabelled} we present some examples from the unlabelled training dataset containing model-agnostic and model-aware instances. Regarding the machine translation (MT) task, we could detect a variety of languages, including Russian, Arabic, Chinese, Yorùbá, Telugu, Tsonga, Uzbek, Sinhalese, Quechuan, Mizo and others. Language information was not provided, so we manually explored the \textit{src} samples in terms of linguistic variability.

Model-agnostic definition modeling (DM) hypotheses contain some `qualifiers', which may guide a model under usage to return a more suitable definition. For example, in the context of the hypothesis containing the definition "(obsolete) An odour," the term "obsolete" indicates that the provided definition is no longer in common use or is outdated. The word "obsolete" is used as a qualifier to convey that the term or concept being defined, in this case, "An odour," was once used to represent a specific meaning but is no longer considered current or applicable in contemporary language.

Another notable observation is that model-aware paraphrase-generation (PG) does not contain any information in \textit{tgt}.
\begin{table*}[h!]
\small
\centering
\begin{tabular}{cp{12cm}}
\hline
\multicolumn{2}{c}{\textbf{Model-agnostic}} \\
\hline
Machine Translation & 'hyp': "Don't worry, it's only temporary.", 'tgt': "Don't worry. It's only temporary.", 'src': \foreignlanguage{russian}{'Не волнуйся. Это только временно.'}, 'ref': 'either', 'task': 'MT', 'model': '' \\
\hline
Definition modelling & 'hyp': '(uncountable) The quality of being oronymy; the state of being oronymy.', 'tgt': 'The nomenclature of mountains, hills and other geographic rises.', 'src': 'An ancient survival in Turkish <define> oronymy </define> is quite possible , but I have not found Nihan Dag on the relevant sheets of the 1 : 200,000 map of Turkey , which are very detailed in matters of oronymy ;', 'ref': 'tgt', 'task': 'DM', 'model': '' \\
\hline
Definition modelling & 'hyp': '(intransitive, obsolete) To make a magazin of; to compose a magazin.', 'tgt': '(colloquial) The act of editing or writing for a magazine.', 'src': "Thus , though Byron is gone after his Don Juan — Scott and Southey out of the rhyme department — Wordsworth stamp - mastering — Coleridge 's poetry in abeyance — Crabbe mute as a fish - Campbell and Wilsont merely <define> magazining </define>", 'ref': 'tgt', 'task': 'DM', 'model': '' \\
\hline
Paraphrase Generation& 'hyp': 'You got something for me, huh?', 'tgt': '', 'src': 'Got something for me?', 'ref': 'src', 'task': 'PG', 'model': '' \\
\hline
\multicolumn{2}{c}{\textbf{Model-aware}} \\
\hline
Machine Translation & 'hyp': "It's like pushing a heavy wheel up a mountain. It splits the nucleus again and releases some energy.", 'tgt': 'Sort of like rolling a heavy cart up a hill. Splitting the nucleus up again then releases some of that energy.', 'src': \begin{CJK}{UTF8}{gbsn}'有點像把沉重的手推車推上山。再次分裂核子然後釋放一些能量\end{CJK}', 'ref': 'either', 'task': 'MT', 'model': 'facebook/nllb-200-distilled-600M' \\
\hline
Machine Translation& 'hyp': 'Our Mailoamiris of the System of Treatment of Ulilaes have created a place for these little ones.', 'tgt': 'We perceive the Foster Care System to be a safety zone for these children.', 'src': 'Maamiris tayo a ti Sistema iti Panangtaripato kadagiti Ulila ket natalged a lugar para kadagitoy nga ubbing.', 'ref': 'either', 'task': 'MT', 'model': 'facebook/nllb-200-distilled-600M'\\
\hline
Definition modeling & 'hyp': 'To be obsequiously interested in .', 'tgt': '( usually followed by over or after ) To fuss over something adoringly ; to be infatuated with someone .', 'src': "Sarah mooned over sam 's photograph for months . What is the meaning of moon ?", 'ref': 'tgt', 'task': 'DM', 'model': 'ltg/flan-t5-definition-en-base' \\
\hline
Paraphrase Generation & {'hyp': "Mr Barros Moura's report looks to the future in my opinion.", 'tgt': '', 'src': 'In my opinion, the most important element of the report by Mr Barros Moura is that it looks to the future.', 'ref': 'src', 'task': 'PG', 'model': 'tuner007/pegasus\_paraphrase'} \\
\hline
\\
\end{tabular}
\caption{
Examples from the unlabelled training set.}
\label{tab:examples-unlabelled}
\end{table*}

\section{Exploratory data analysis}
\label{sec:analysis}

\begin{figure}[h!]
    \centering
    \includegraphics[width=\columnwidth]{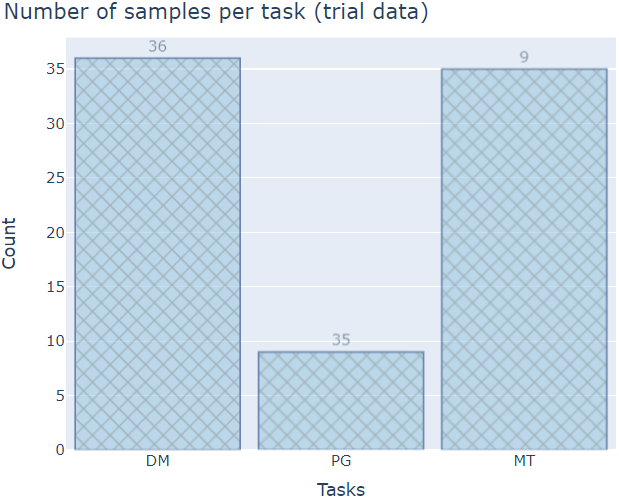}
    \caption{Distribution of per task samples in the initially released trial set.}
    \label{fig:trial-data}
\end{figure}

\begin{figure*}[h!]
  \centering
  \begin{subfigure}{0.48\textwidth}
    \includegraphics[width=\linewidth]{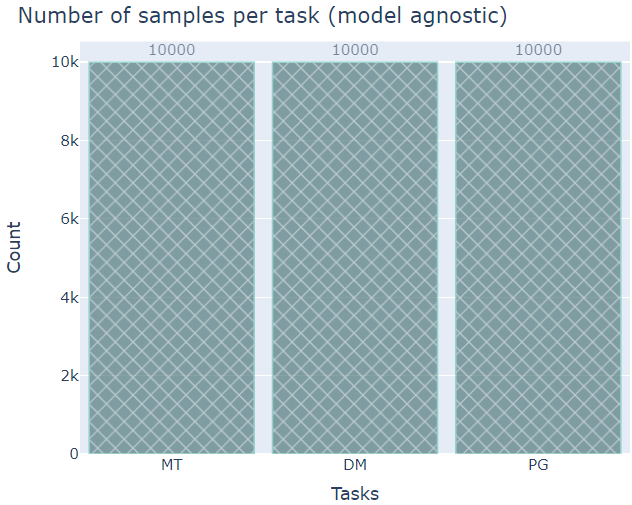}
    \caption{Model-agnostic sample distribution in the training set.}
  \end{subfigure}
  \hfill
  \begin{subfigure}{0.48\textwidth}
    \includegraphics[width=\linewidth]{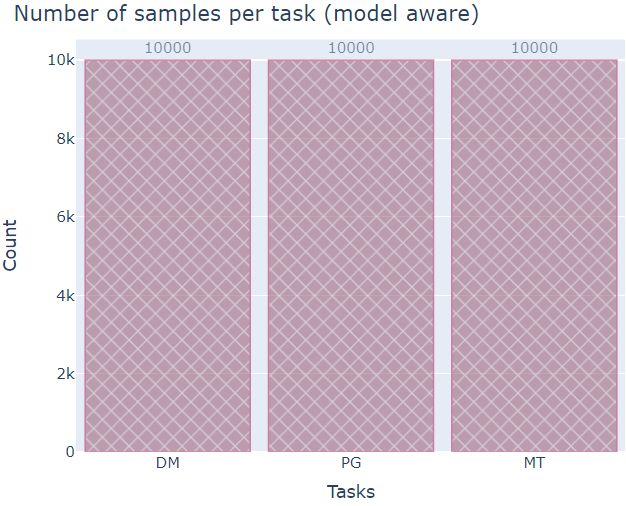}
    \caption{Model-aware sample distribution in the training set.}
  \end{subfigure}
  \caption{Distribution of unlabelled training samples per task in both model-agnostic and model-aware settings.}
  \label{fig:training-distr}
\end{figure*}

\paragraph{Trial set}
We explore the frequency of each task occurring within samples from different dataset splits, commencing from the initially released trial set.  In Figure \ref{fig:trial-data} we present the task distribution of the first 80 trial samples.
\paragraph{Unlabelled data (training set)}
Figure \ref{fig:training-distr} represents the distribution in the training set. In both model-agnostic and model-aware settings each task contains an equal number of samples (10k samples per task in each setting). In our methodologies, we abstained from utilizing the provided unlabeled training dataset as it did not align with our main approaches.

\begin{figure*}[h!]
  \centering
  \begin{subfigure}{0.48\textwidth}
    \includegraphics[width=\linewidth]{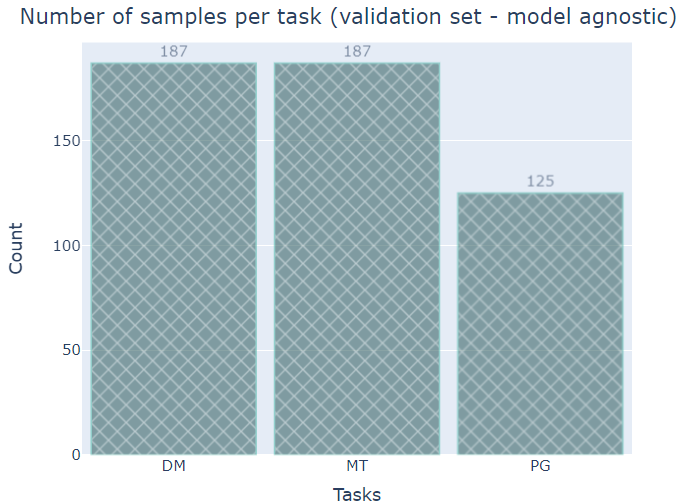}
    \caption{Model-agnostic sample distribution in the validation set.}
  \end{subfigure}
  \hfill
  \begin{subfigure}{0.48\textwidth}
    \includegraphics[width=\linewidth]{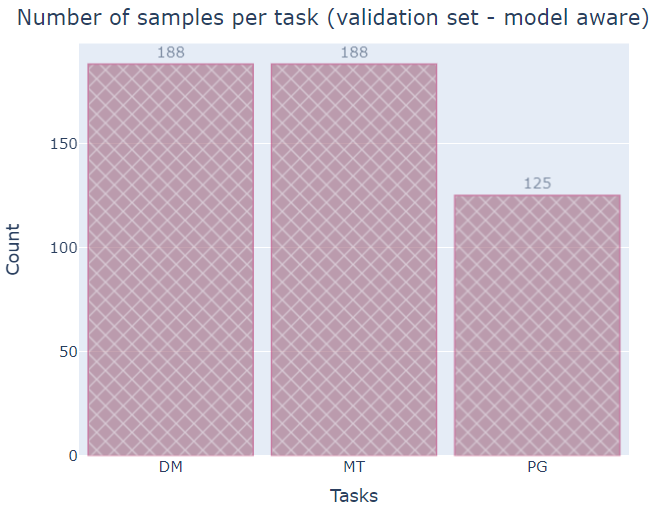}
    \caption{Model-aware sample distribution in the validation set.}
  \end{subfigure}
  \caption{Distribution of labeled validation samples per task in both model-agnostic and model-aware settings.}
  \label{fig:val-distr}
\end{figure*}

\begin{figure*}[h!]
  \centering
  \begin{subfigure}{0.48\textwidth}
    \includegraphics[width=\linewidth]{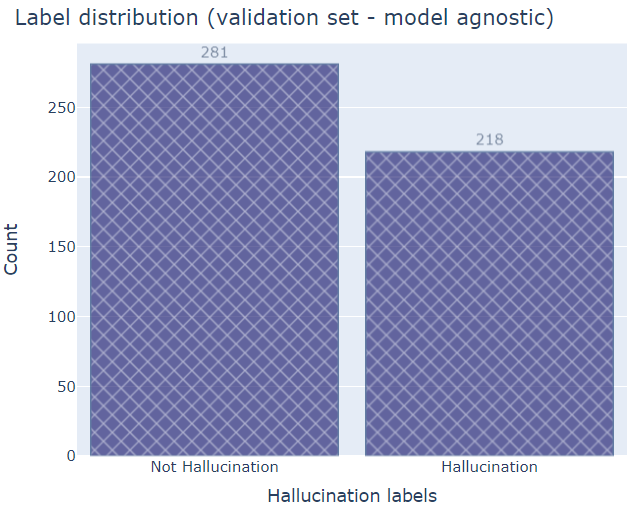}
    \caption{Model-agnostic label distribution in the validation set.}
  \end{subfigure}
  \hfill
  \begin{subfigure}{0.48\textwidth}
    \includegraphics[width=\linewidth]{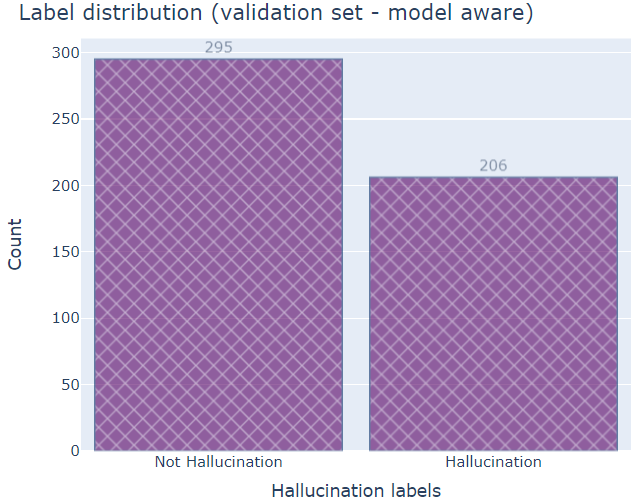}
    \caption{Model-aware label distribution in the validation set.}
  \end{subfigure}
  \caption{Distribution of validation labels in both model-agnostic and model-aware settings.}
  \label{fig:val-labels}
\end{figure*}

\begin{figure*}[h!]
  \centering
  \begin{subfigure}{0.48\textwidth}
    \includegraphics[width=\linewidth]{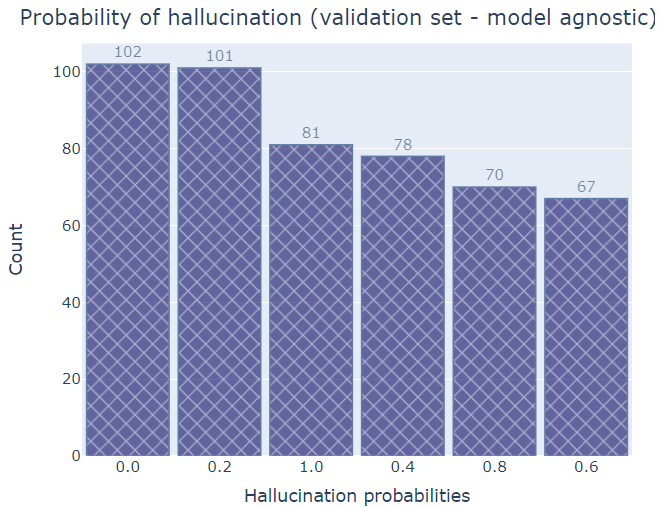}
    \caption{Model-agnostic hallucination probability distribution in the validation set.}
  \end{subfigure}
  \hfill
  \begin{subfigure}{0.48\textwidth}
    \includegraphics[width=\linewidth]{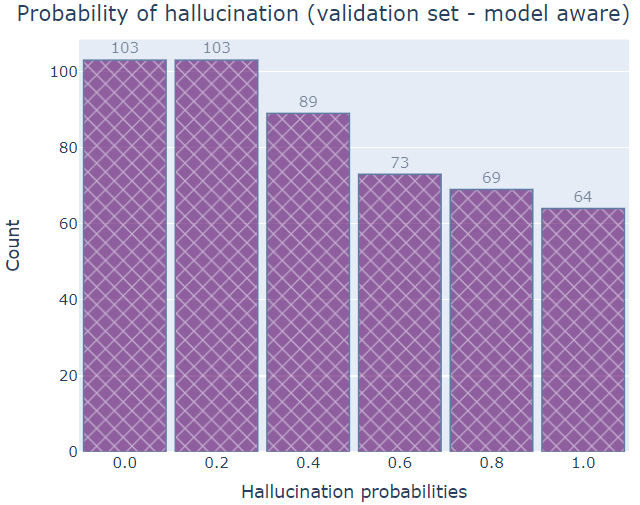}
    \caption{Model-aware hallucination probability distribution in the validation set.}
  \end{subfigure}
  \caption{Distribution of hallucination probability (majority voting among human annotators' labeling) in both model-agnostic and model-aware settings in the validation set.}
  \label{fig:val-phal}
\end{figure*}

\begin{figure*}[h!]
  \centering
  \begin{subfigure}{0.48\textwidth}
    \includegraphics[width=\linewidth]{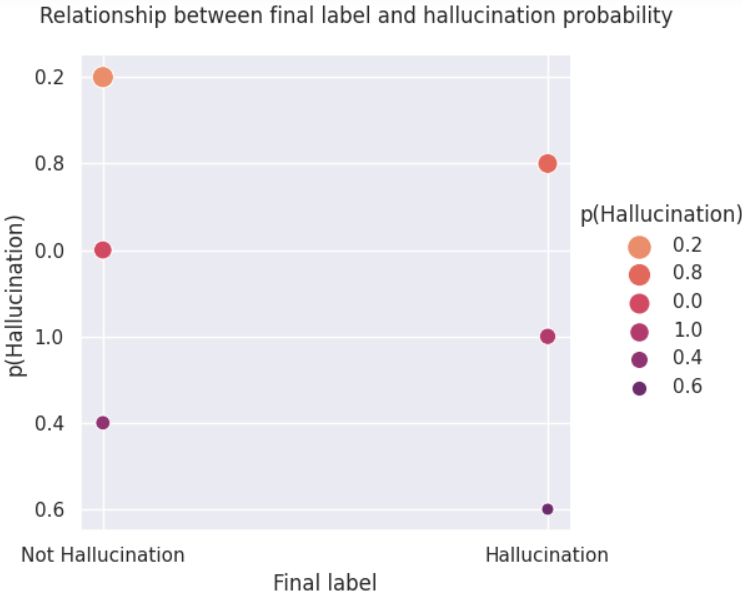}
    \caption{Hallucination probability per label (Model-agnostic).}
  \end{subfigure}
  \hfill
  \begin{subfigure}{0.48\textwidth}
    \includegraphics[width=\linewidth]{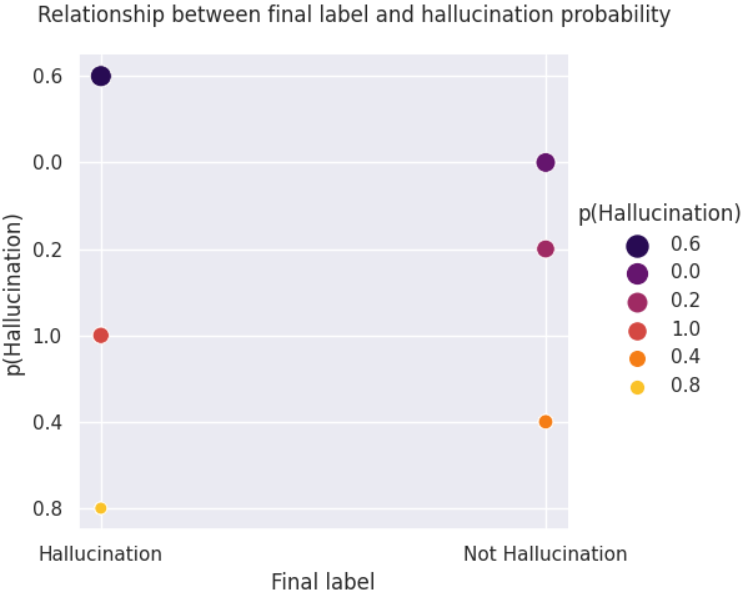}
    \caption{Hallucination probability per label (Model aware).}
  \end{subfigure}
  \caption{Distribution of hallucination probability in each validation label ('Hallucination' vs 'Not Hallucination'). Annotators significantly agree on whether a sample contains a hallucination or not.}
  \label{fig:relplot-val}
\end{figure*}

\begin{figure*}[h!]
  \centering
  \begin{subfigure}{0.48\textwidth}
    \includegraphics[width=\linewidth]{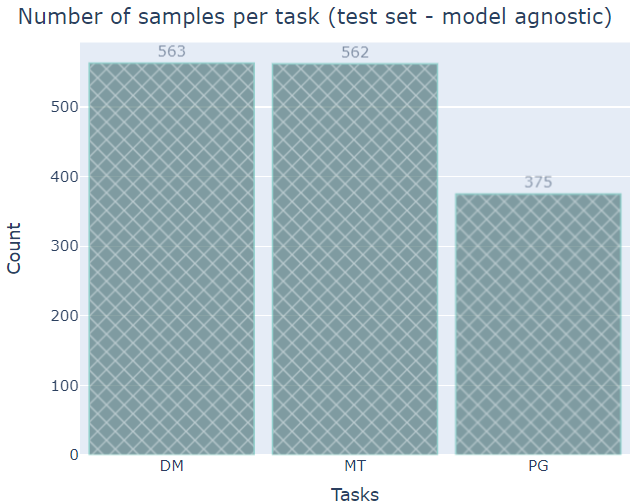}
    \caption{Model-agnostic sample distribution in the test set.}
  \end{subfigure}
  \hfill
  \begin{subfigure}{0.48\textwidth}
    \includegraphics[width=\linewidth]{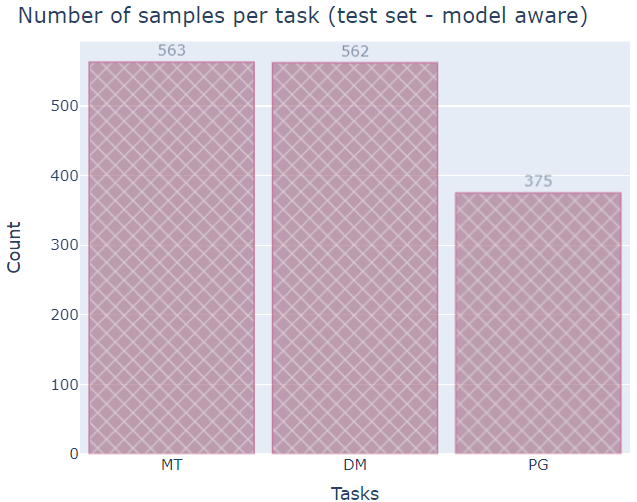}
    \caption{Model-aware sample distribution in the test set.}
  \end{subfigure}
  \caption{Distribution of labeled test samples per task in both model-agnostic and model-aware settings.}
  \label{fig:test-distr}
\end{figure*}

\begin{figure*}[h!]
  \centering
  \begin{subfigure}{0.48\textwidth}
    \includegraphics[width=\linewidth]{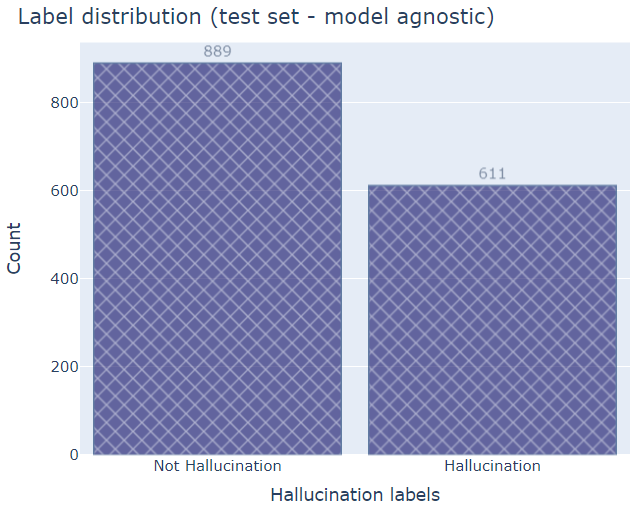}
    \caption{Model-agnostic label distribution in the test set.}
  \end{subfigure}
  \hfill
  \begin{subfigure}{0.48\textwidth}
    \includegraphics[width=\linewidth]{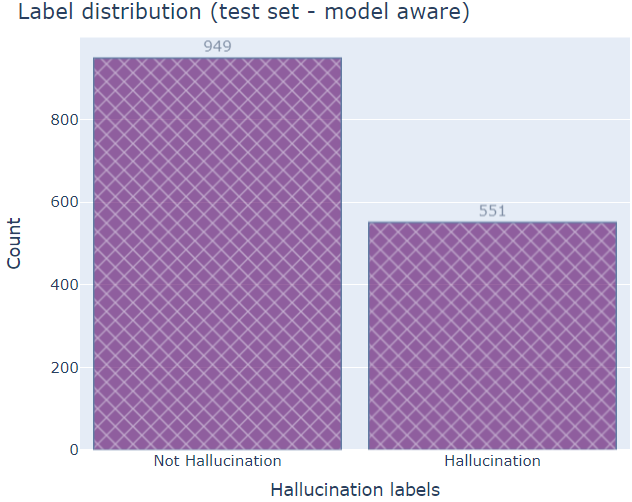}
    \caption{Model-aware label distribution in the test set.}
  \end{subfigure}
  \caption{Distribution of test labels in both model-agnostic and model-aware settings.}
  \label{fig:test-labels}
\end{figure*}

\begin{figure*}[h!]
  \centering
  \begin{subfigure}{0.48\textwidth}
    \includegraphics[width=\linewidth]{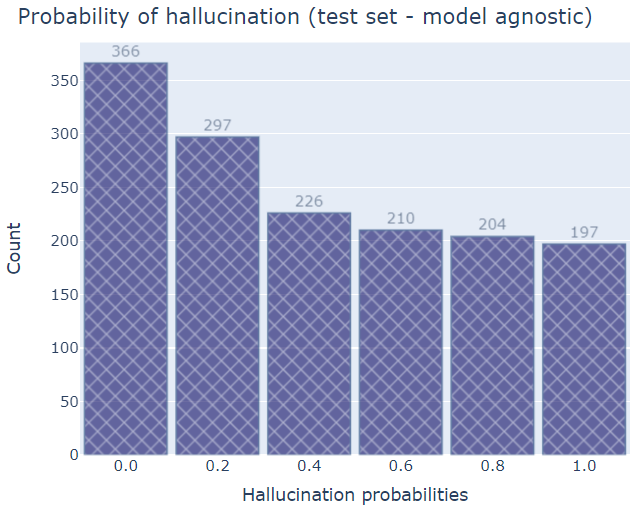}
    \caption{Model-agnostic hallucination probability distribution in the test set.}
  \end{subfigure}
  \hfill
  \begin{subfigure}{0.48\textwidth}
    \includegraphics[width=\linewidth]{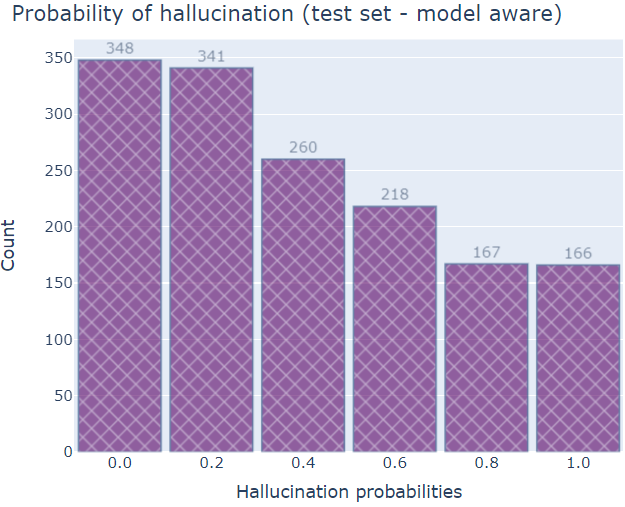}
    \caption{Model-aware hallucination probability distribution in the test set.}
  \end{subfigure}
  \caption{Distribution of hallucination probability (majority voting among human annotators' labeling) in both model-agnostic and model-aware settings in the test set.}
  \label{fig:test-phal}
\end{figure*}

\begin{figure*}[h!]
  \centering
  \begin{subfigure}{0.48\textwidth}
    \includegraphics[width=\linewidth]{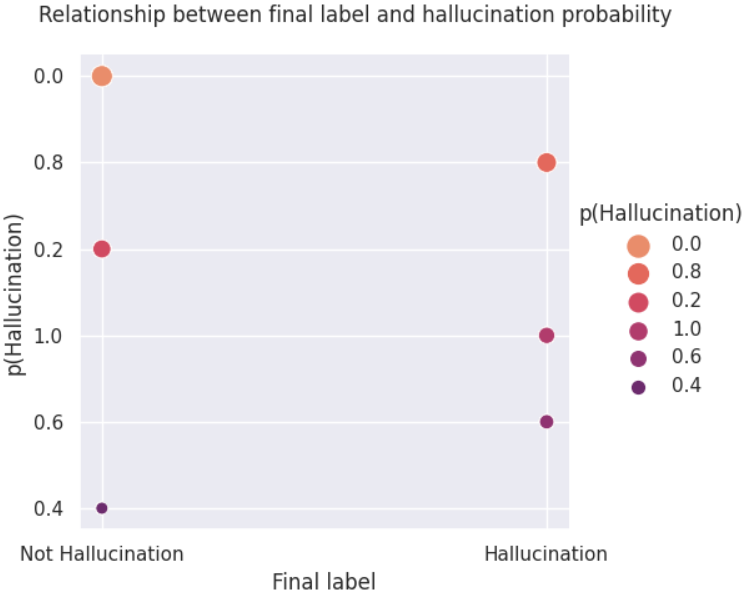}
    \caption{Hallucination probability per label (Model-agnostic).}
  \end{subfigure}
  \hfill
  \begin{subfigure}{0.48\textwidth}
    \includegraphics[width=\linewidth]{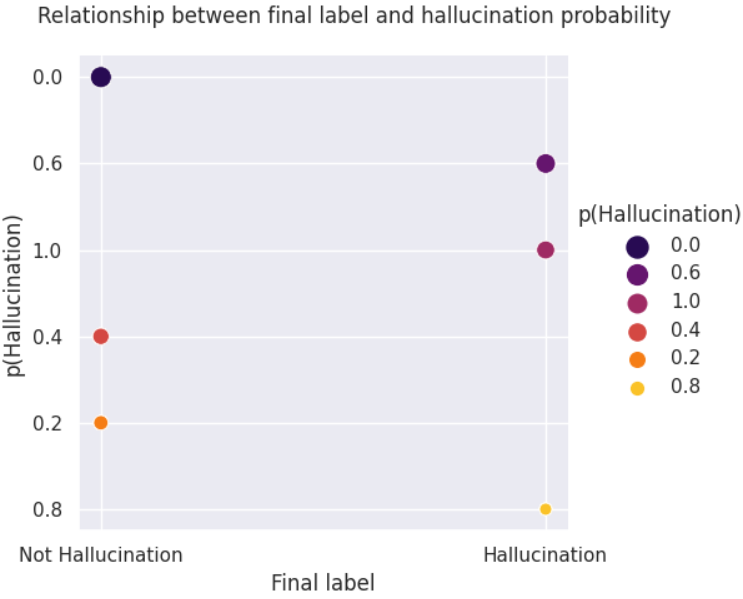}
    \caption{Hallucination probability per label (Model aware).}
  \end{subfigure}
  \caption{Distribution of hallucination probability in each test label ('Hallucination' vs 'Not Hallucination). Annotators significantly agree on whether a sample contains a hallucination or not.}
  \label{fig:relplot}
\end{figure*}

\paragraph{Validation set}
Moving on to labeled data, we commence with the validation (dev) set, for which we present per task distributions in Figure \ref{fig:val-distr}. We observe a difference in the distribution of labels in comparison to the balanced training set distribution of Figure \ref{fig:training-distr}; nevertheless, since we do not exploit any unlabelled instance, this does not pose a limitation for us at this point.

We proceed with studying the validation set label distribution. Related results are presented in Figure \ref{fig:val-labels}, denoting label imbalance in both model-agnostic and model-aware settings.

The distribution of hallucination probability is presented in Figure \ref{fig:val-phal}. As expected, low p('Hallucination') values are more common (indicating that fewer annotations voted for the presence of a hallucinatory instance), since 'Not Hallucination' is the majority label in both settings. Ideally, we wish borderline probabilities to be low:
The highest the disagreement for a certain sample, the closest to the 0.5 threshold the hallucination probability will be (a p('Hallucination')=0.4 denotes that 3/5 annotators voted for 'Not Hallucination', while the rest 2/5 voted for the opposite; on the other hand, a p('Hallucination')=0.6 denotes that 3/5 annotators voted for 'Hallucination', while the rest 2/5 voted for 'Not Hallucination'. Therefore, the highest uncertainty is observed close to the 0.5 boundary). This requirement is adequately satisfied especially in the model-agnostic case (left plot of Figure \ref{fig:val-phal}), where p('Hallucination')=0.6 is the least frequent.

Further insights can be obtained by looking at Figure \ref{fig:relplot-val}: when smaller dots are assigned to probabilities close to the 0.5 threshold, the annotators' disagreement is lower, therefore classifying a sample as 'Hallucination' or 'Not hallucination' is less uncertain. Indeed, the less frequently appearing p('Hallucination')=0.4 and p('Hallucination')=0.6 values in the model-agnostic case denote high separability between hallucinated and non-hallucinated samples. However, highly certain values, such as p('Hallucination')=0.0 and p('Hallucination')=1.0 only rank in the middle, therefore even if samples are separable with low uncertainty, some minor disagreement persists (1/5 annotators frequently disagrees with the rest). Overall, annotators are almost equally confident in classifying 'Hallucination' and 'Not Hallucination' samples, as indicated by the matching pattern regarding label uncertainty for both labels.
The model-aware case is more confusing, with p('Hallucination')=0.6 scoring the highest; therefore, classifying a sample as 'Hallucination' is often accompanied by high uncertainty. On the contrary, uncertainty is lower for the 'Not Hallucination' label, with p('Hallucination')=0.0 ranking as the second most frequent probability. We can conclude that in the model-aware setting of the validation set, annotators are more confident in recognizing the 'Not Hallucination' class in comparison to the 'Hallucination' one.

\paragraph{Test set}

As for the test set, Figure \ref{fig:test-distr} represents the number of samples per task for both settings. Note that the test task distribution is similar to the validation distribution of Figure \ref{fig:val-distr}with PG being a minority label in all cases.  

In terms of ground-truth label (Hallucination vs Not Hallucination), Figure \ref{fig:test-labels} highlights some label imbalance, rendering the prediction of 'Not Hallucination' more possible in a random setup for both model-agnostic and model-aware settings. This label distribution matches the validation set label distribution (Figure \ref{fig:val-labels}), for which 'Not Hallucination' was the majority class as well.

Hallucination probability per setting is depicted in Figure \ref{fig:test-phal}, with lower hallucination values in the range [0, 0.2) being more common. This is again somehow expected since 'Not Hallucination' is the majority class in test labels. More insights can be obtained by looking at Figure \ref{fig:relplot}, which relates the hallucination probability with the label. 
Especially in the model-agnostic setting (Figure \ref{fig:relplot} - left), the p('Hallucination')=0.4 and p('Hallucination')=0.6 values are the lowest (smaller dots), while p('Hallucination')=0.0 is the highest, denoting that annotators are often certain regarding non-hallucinated samples. Certainty for hallucinated samples is somehow lower, as p('Hallucination')=1.0 lies somewhere in the middle. Nevertheless, p('Hallucination')=0.8 is the second more frequent value denoting that 4/5 annotators frequently annotate a sample as 'Hallucination'. By observing the right plot of Figure \ref{fig:relplot}, we conclude that certainty is lower in the model-aware setting. Even though p('Hallucination')=0.0 remains the most frequent probability, indicating high agreement regarding non-hallucinated samples, the p('Hallucination')=0.6 value stands in the second place. Therefore, many samples classified as 'Hallucination' achieved this label with low agreement (3/5 annotators). Also, the p('Hallucination')=0.2 and p('Hallucination')=0.8 are the lowest, denoting that higher agreement (4/5 annotators agreeing) is rare for both 'Hallucination' and 'Not Hallucination' labels. We can assume that model-aware samples are harder by nature to be classified in any of the labels.
\section{NLI-Hyperparameters}
The hyperparameters utilized for the NLI model fine-tuning mirrored those employed during the training of the initial model. The selection of hyperparameters followed a series of experiments, which yielded significantly lower levels of accuracy. Some of the experiments are displayed in the Table \ref{Experiments for hyperparameters}
\tabcolsep=0.1cm
\begin{table}[h!]
\centering
\small
\begin{tabular}{ccccc}
\hline
\textbf{epochs} & \textbf{lr} & \textbf{warmup ratio} 
&\textbf{weight decay} &\textbf{accuracy} \\
\hline
5 & 2e-05 & 0.06 & 0.01 & 0.83\\
10 & 2e-06 & 0.1 & 0.01 & 0.75\\
5 & 2e-04 & 0.01 & 0.05 & 0.53\\
5 & 2e-05 & 0.05 & 0.001 & 0.8\\
5 & 2e-06 & 0.08 & 0.1 & 0.79\\

\hline
\end{tabular}
\caption{\label{Experiments for hyperparameters}
Accuracy on trial-set from experiments with hyperparameters. The first row displays the hyperparameters chosen for finetuning}
\end{table}
\end{document}